\def\year{2019}\relax
\documentclass[letterpaper]{article} 
\usepackage{aaai19_arxiv}  
\usepackage{times}  
\usepackage{helvet} 
\usepackage{courier}  
\usepackage[hyphens]{url}  
\usepackage{graphicx} 
\usepackage{graphicx}  
\usepackage[utf8]{inputenc} 
\usepackage{url}            
\usepackage{booktabs}       
\usepackage{amsfonts}       
\usepackage{nicefrac}       
\usepackage{microtype}      

\usepackage{latexsym}
\usepackage{amsmath}
\usepackage{amsfonts}
\usepackage{mathtools}
\usepackage{url}
\usepackage{graphicx}
\usepackage{appendix}
\usepackage{multirow}
\usepackage{color}
\usepackage[table,xcdraw,dvipsnames]{xcolor}

\definecolor{lightblue}{HTML}{E0ECF7}
\definecolor{darkblue}{HTML}{092E6B}
\newcommand{\win}[1]{{\colorbox{lightblue}{\sf #1}}}
\newcommand{\lose}[1]{{\colorbox{darkblue}{\sf \color{white}{#1}}}}

\newcommand{\citet}[1]{\citeauthor{#1} \shortcite{#1}}
\newcommand{\citep}{\cite}

\title{{\sc Acute-eval}: Improved dialogue evaluation  with optimized questions and
multi-turn comparisons}

\author{
  Margaret Li\\
  Facebook AI Research \\
  {\tt \small{margaretli@fb.com}} \\\And
  Jason Weston \\
  Facebook AI Research \\
  {\tt \small{jase@fb.com}} \\\And
  Stephen Roller \\
  Facebook AI Research \\
  {\tt \small{roller@fb.com}}
}

\begin{document}

\maketitle

\begin{abstract}
While dialogue remains an important end-goal of natural language research, the difficulty of evaluation is an oft-quoted reason why it remains troublesome to make real progress towards its solution. Evaluation difficulties are actually two-fold: not only do automatic metrics not correlate well with human judgments, but also human judgments themselves are in fact difficult to measure. The two most used human judgment tests, single-turn pairwise evaluation and multi-turn Likert scores, both have serious flaws as we discuss in this work.

We instead provide a novel procedure involving comparing two full dialogues, where a human judge is asked to pay attention to only one speaker within each, and make a pairwise judgment. The questions themselves are optimized to maximize the robustness of judgments across different annotators, resulting in better tests. We also show how these tests work in self-play model chat setups, resulting in faster, cheaper tests. We hope these tests become the de facto standard, and will release open-source code to that end.
\end{abstract}

\section{Introduction}
\label{sec:intro}

\begin{figure}[t]
\centering
\includegraphics[width=0.47\textwidth]{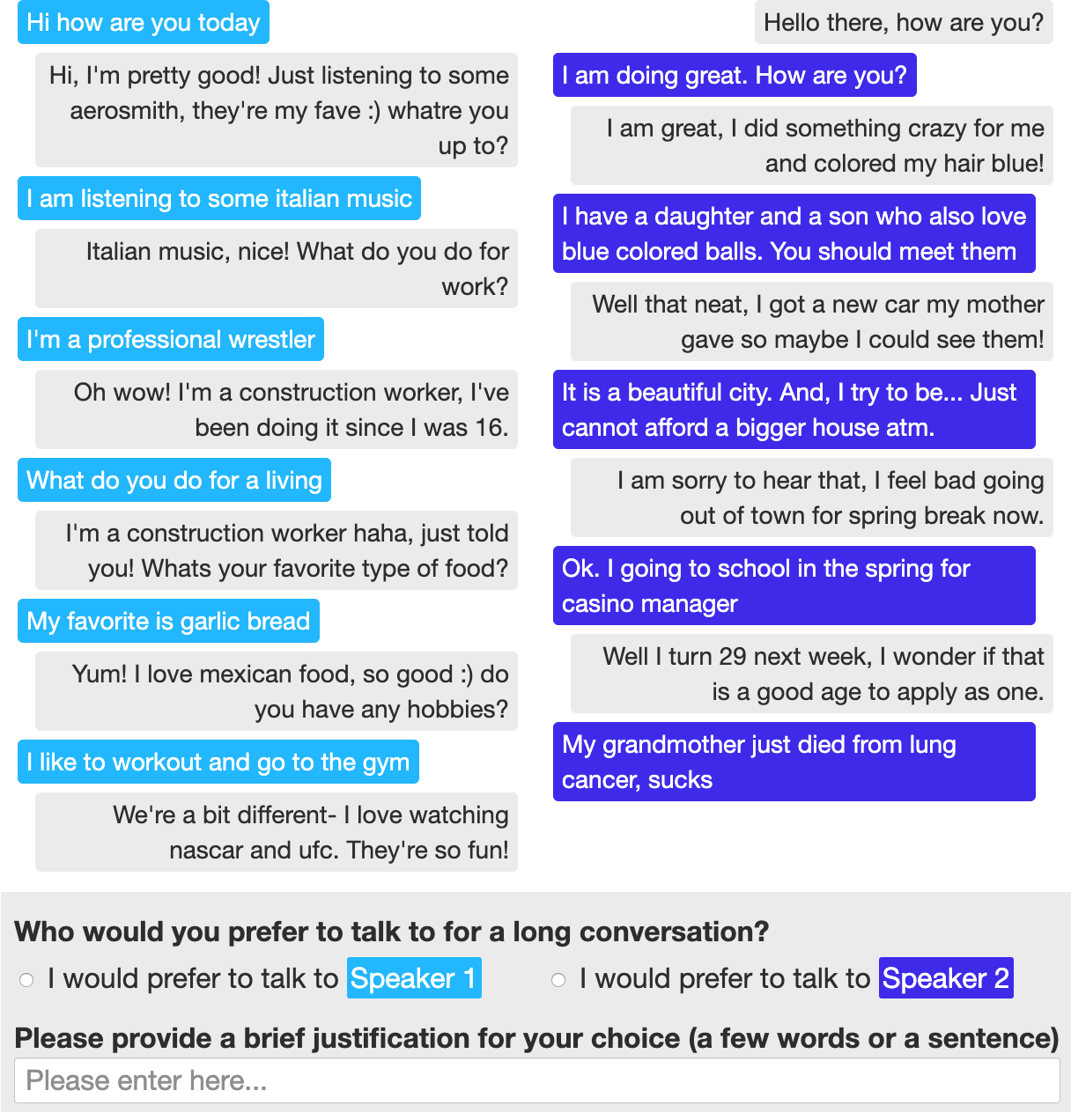}
\caption{{\sc Acute-eval} asks humans to compare two multi-turn dialogues, and independent of the gray speakers, choose between Speaker 1 (light blue) and Speaker 2 (dark blue).}
\label{fig:ab_ui1}
\end{figure}

Dialogue between human and machine is an important end-goal of natural language research. The open-ended nature of generating sequences in a multi-turn setup naturally makes the task difficult to evaluate -- with full evaluation possessing many of the difficulties of the task itself as it requires deep understanding of the content of the conversation.
As in many other natural language generation (NLG) tasks, automatic metrics have not been shown to have a clear correlation with human evaluations \cite{liu2016not,lowe2017towards}. This means the current standard for all dialogue research involves human trials, which slows down research and greatly increases the cost of model development.

Unfortunately, human judgments are themselves difficult to measure. The two most used approaches, single-turn pairwise evaluation \citep{vinyals2015neural,Li2016AModel}, and multi-turn Likert scores \citep{venkatesh2018evaluating,Zhang2018PersonalizingToo,see2019goodconversation,dinan2018wizard,dinan2019second} have serious limitations. Single-turn pairwise evaluation provides the benefits and simplicity of an A/B test, 
allowing for cheap and fast annotations, with comparisons that are robust to annotator score bias, but fail to take into account the multi-turn aspect of conversations. To give a trivial example,  such comparisons fail to capture whether the model would repeat itself in a multi-turn conversation because they only look at one turn; repetition is a known issue that humans dislike \citep{see2019goodconversation}.

Multi-turn Likert scores require the annotator to have a multi-turn conversation and then provide an integer score, which is more costly and time-consuming to run but evaluates full conversations more accurately.
The integer scores however suffer from differing bias and variance per 
annotator, which researchers have tried to mitigate \citep{kulikov2018importance},
but nevertheless due to its lack of sensitivity 
often yields comparisons that are not statistically significant. Furthermore, due to strong anchoring effects during model evaluation, i.e. that annotators are affected by the first systems they evaluate, Likert comparisons are generally not comparable across multiple papers. This mandates that evaluations of new models be simultaneously collected with baselines, further increasing the cost of developing additional models \cite{see2019goodconversation}.

In this work we introduce {\sc Acute-eval}, a method that combines the benefits, and attempts
to mitigate the deficiencies,
of the above two approaches by introducing a pairwise relative comparison setup for multi-turn dialogues. In each trial, we show the annotator two whole conversations,  with the second speaker in each conversation highlighted, as the judgment should be independent of the quality of the first speaker, see Figure~\ref{fig:ab_ui1}. 
We then show a carefully worded question with two choices: speaker A or B, where the question measures a desired quality such as which speaker is more engaging, interesting or knowledgeable. 
Our experiments show that annotators perform well in this setup,
and that our method can reveal subtle but significant differences between conversational models that other approaches, such as multi-turn Likert, cannot.

Overall, our work provides the following contributions: 
\begin{itemize}
    \item A new evaluation method with a clear mechanism that provides fast, cheap iteration. This evaluation method allows efficient reuse of data from prior papers, allowing new models to be evaluated independently of baselines, and dramatically lowers the cost of annotation.
    \item We optimize question choices to find those with the highest agreement, increasing confidence in the desired test. We provide the wording of the questions that we found to work best for several questions of interest (most engaging, human, interesting or knowledgeable conversationalist) for further research use. 
    \item We provide an explicit benchmark comparison between current best performing retrieval and generative models on two recent tasks, PersonaChat \citep{Zhang2018PersonalizingToo} and Wizard of Wikipedia \citep{dinan2018wizard} for several question choices, revealing the current state-of-the-art, and to be used for benchmarking on these tasks in the future.
    \item We show that our test can be applied to self-chats rather than human-model conversation logs, which can reveal problems with existing models at a cheaper price, and provides high agreement with the human-model evaluations.
    \item We will release the code for running these tests. 
\end{itemize}

\section{Related Work}  
\label{sec:rel_work}
\label{subsec:rel_work_dialog}

Dialogue tasks have traditionally been separated into two areas: goal-oriented and chitchat. Goal-oriented tasks typically have a clearer evaluation,
e.g. task completion can be measured if the correct actions are taken \citep{lemon,henderson2014second,bordes2016learning,asri2017frames,wen2016network}. Chitchat tasks are more open ended, and instead feature 
conversations without a 
precise goal that can be automatically evaluated. For example, conversations 
where two speaking partners are discussing interests \citep{Zhang2018PersonalizingToo} or topics \citep{dinan2018wizard}.
We study the latter in this work.

Evaluation of chitchat tasks with automatic metrics is difficult precisely because of their open-ended nature. For example, the answer to the question ``What are you doing tonight?'' has many possible answers, each with little word overlap. This means standard metrics for tasks like question-answering or machine translation do not work well, and have poor correlation with human judgments \citep{liu2016not,novikova2017we}.
Nevertheless, a number of studies do report automatic metrics, 
without human studies
\citep{serban2016building,parthasarathi2018extending}.
Researchers have made attempts to improve automatic evaluation, 
trying methods such  as adversarial evaluation \citep{li2017adversarial}, learning a scoring model 
\citep{lowe2017towards}, 
or a learnt ensemble of automatic metrics \citep{DBLP:journals/corr/abs-1906-09308},
but their value is as yet not fully understood.

Currently the standard approach in chitchat dialogue is to perform human evaluations  \citep{vinyals2015neural,li2015diversity,li2016deep,venkatesh2018evaluating,Zhang2018PersonalizingToo,dinan2018wizard},
typically reporting a judgment 
such as conversation quality or appropriateness 
via a Likert scale or pairwise comparison.
While conversations are naturally multi-turn, pairwise setups typically
consider single turn evaluations, taking the ``gold'' dialogue history from human-human logs, and only consider altering a single utterance. A more complete multi-turn evaluation is typically measured with a Likert scale (usually 1-4 or 1-5) after the conversation takes place.
Some works such as \cite{see2019goodconversation} ask a series of questions relating to different aspects of conversational ability.
There are some notable variants from these standard setups.
\citet{novikova2018rankme} provide a method that combines continuous scales and relative assessments, but in single-turn, rather than multi-turn evaluation.
\citet{DBLP:journals/corr/abs-1906-09308} compare human evaluations to
automatic metrics computed on self-chats. Note that we also use self-chats in this work, but we evaluate these with humans, rather than automatic metrics.

Finally, this work expands upon some of the ideas present in \citet{see2019goodconversation}. In that work, a test for interestingness  of a specificity-controlled model conducted with pairwise chat logs was mentioned, similar to the ones used here, but was not the focus of their work. 
In our work, we conduct a full study of novel variants of this approach,
consider optimizing the questions for robust measurements over four types of questions, 
utilize self-chat logs
in addition to human-bot logs,
and benchmark state-of-the-art models across two recent tasks.

\section{Method: {\sc Acute-eval}}

To compare two dialogue models, model A and model B, our evaluation 
 asks humans to directly compare side-by-side multi-turn dialogues 
conducted by these models.
See Figure~\ref{fig:ab_ui1} for an example. 

Our method is thus the following: (1) collect conversation logs for model A; similarly for model B.  (2) In a number of trials, ask annotators 
to make binary judgments between sampled pairs from the logs, 
and collate the results to determine the winner, either A or B, and the statistical significance.

We consider different approaches to step (1) and (2) below.

\paragraph{Human-Model chats}

Our standard setup is to compare
conversation logs between models and humans. 
In each evaluation trial we then show a human annotator two of the 
previously obtained conversations, one of model $A$ conversing with a human, and one of model $B$ conversing with a (possibly different) human.
 The annotator sees the conversations side by side on the same screen, with the two models' utterances highlighted in different colors, and the human utterances in gray to minimally distract from the models.  
 
 The annotator is posed a question phrasing (e.g. ``which speaker is more knowledgeable'' or ``which speaker sounds more human?''), and asked to make a binary choice between model $A$ and model $B$. They are strongly encouraged to provide a short text justification for their choice. 
We collect $N$ trials of such pairwise judgments, and use them to decide which model wins.  Statistical significance can be computed using a binomial test.

\paragraph{Self-Chats}
Human-model conversation logs are themselves time-consuming and expensive to
collect, which limits rapid iterative model development.
We investigate if it is possible to remove the human from the conversation, and only use human annotators in the final pairwise conversation evaluation step. The concept of self-chats
\cite{li2016deep,DBLP:journals/corr/abs-1906-09308}, whereby a model talks to itself, playing the roles of both speaking partners, has been previously explored in other contexts. Such logs are easy
to collect for models A and B, involving simply running inference for both speaker roles.
We then use these logs in the {\sc Acute-eval} pairwise comparison setup as described above.

\paragraph{Question Optimization}

So far, we have not detailed the actual question(s) asked of the annotators.
The framing and phrasing of questions in surveys is known to greatly affect the direction of responses, and therefore, in the case of evaluation, inter-annotator agreement. Though this has been noted in prior work \citep{lowe2017towards}, we have found no systematic experimentation on question formulation or task presentation. We therefore aim to propose and evaluate multiple potential question wordings to achieve higher agreement.

To do this, we build an initial test that compares human-human logs with human-model logs where the model is a relatively low quality baseline model. The aim is that there should be a clear and agreeable difference between human and model which is visible to human annotators. We ask annotators to make judgments between these two, where we choose pairs where the human should be judged as superior.

We then run independent trials with different question phrasing, and find the questions with highest inter-annotator agreement.
The winning questions can then be used in future experiments by ourselves, and other researchers. Although having high inter-annotator agreement does not guarantee that crowdworkers interpret the question as intended, it increases the chance the question is understood uniformly. 
That is, the researcher still has to exercise care in the formulation of the question so that they believe it measures the quantity they are interested in. 
In our experiments we find  questions with high-agreement rate over four axes: engagingness, interestingness, knowledge and humanness.

\paragraph{Annotation Quality}

We use crowdworkers for our annotations. We recommend limiting the number of annotations a single worker may complete to be only a few pairs (in our experiments, if we are making $N$ model comparisons then we allow $N$ annotations). In preliminary trials, we found that limiting the influence of any one worker was important for replicability, but that results were highly consistent across multiple runs with this limitation.

Additionally, the first comparison any worker is asked to annotate consists of a conversation between a weak baseline model and human, and a human-human conversation. If a worker fails to rate the human-human conversation as better, we remove their annotations from the results, in order to remove poor quality annotators. We additionally remove workers who never give a reason for their choice. Note that adding such worker quality tests to pairwise annotation tasks is straightforward where the gold annotation is known, while it is harder for Likert tests which have integer scores. One may also increase the number of quality-control annotations to decrease the likelihood of fraudulent workers, but we found using a single control question had a reasonable cost-noise ratio.

Each specific \textit{pair} of conversations is shown at most once, given that there are at least as many possible pairs of conversations as desired annotations. If there are more conversations available for each model than desired annotations, each \textit{conversation} is shown at most once - that is, in only one annotation. We found that maximizing the diversity of pairs improved robustness of our evaluation across multiple replication experiments.

\section{Experiments}

\begin{table*}[!ht]
    \centering
    \begin{small}
    \begin{tabular}{p{8cm}p{5.6cm}r}
        \toprule
        {\bf Question} & {\bf Choice 1} &  {\bf Agrm.} \\
        \midrule
        \multicolumn{3}{c}{Engagingness (PersonaChat)}\\
        \cmidrule{1-3}
        Which speaker is more engaging to talk to? &    Speaker 1 is more engaging & 82.5\%\\
        Who would you prefer to talk to for a long conversation? & I would prefer to talk to Speaker 1 & *{\bf 87.5\%}\\
        Which speaker do you think is more captivating? & Speaker 1 is more captivating than Speaker 2 & 84.2\% \\
        \cmidrule{1-3}

        \multicolumn{3}{c}{Interestingness (PersonaChat)}\\
        \cmidrule{1-3}
        If you had to say one of these speakers is interesting and one is boring, who would you say is more interesting? &    Speaker 1 is more interesting & *{\bf 86.7\%} \\
        Which speaker is more interesting to talk to? &   Speaker 1 is more interesting & *81.5\% \\
        Which speaker is more boring to talk to? &  Speaker 1 is more boring & 69.6\% \\
        Who would you rather talk to for fun? & Speaker 1 is more fun & 70.8\%\\

        \cmidrule{1-3}
        \multicolumn{3}{c}{Humanness (PersonaChat)}\\
        \cmidrule{1-3}
        Which speaker sounds more human? &  Speaker 1 sounds more human & *{\bf 76.9}\% \\
        If you had to guess that one speaker is human and one is a bot, which do you think is human? &   Speaker 1 sounds human & 71.4\%\\
        Which speaker sounds more like a real person? & Speaker 1 sounds more like a real person & 76.9\%\\
        
        \cmidrule{1-3}
        \multicolumn{3}{c}{Knowledgeable (Wizard of Wikipedia)}\\
        \cmidrule{1-3}
        Which speaker is more knowledgeable? &  Speaker 1 is more knowledgeable & *88.9\% \\
        If you had to say that one speaker is more knowledgeable and one is more ignorant, who is more knowledgeable? &   Speaker 1 is more knowledgeable & *{\bf 100\%}\\
        Which speaker is more well-informed? &  Speaker 1 is more well-informed & *85.0\%\\
        \bottomrule
    \end{tabular}
    \end{small}
    \caption{{\bf Optimizing questions}: we measure the agreement rates for the most chosen response for different phrasings of questions, and choose the most agreed upon versions.
    Starred agreements indicate statistical significance (binomial test, $p<.05$), and bold
    agreements indicate the question was used in future trials.}
    \label{tab:abtestquestions}
\end{table*}


We perform experiments on two tasks, PersonaChat and Wizard of Wikipedia, which evaluate
different aspects of conversational ability. We first optimize the questions to maximize worker agreement, and then benchmark existing state-of-the-art models on each task.

\subsection{PersonaChat task}
\label{sec:personachat}
PersonaChat \citep{Zhang2018PersonalizingToo} is a chitchat dialogue task involving two participants (two humans or a human and a bot). Each participant is given a \textit{persona} -- a short collection of personal traits such as \textit{I'm left handed} or \textit{My favorite season is spring} -- and are instructed to get to know each other by chatting naturally using their designated personas, for 6--8 turns. The original dataset contains nearly 9000 human-human training conversations; most models are pretrained with a larger corpus, and then fine-tuned on this set.  

PersonaChat was the subject of the NeurIPS 2018 ConvAI2 Challenge \citep{dinan2019second}, in which competitor's models were first evaluated with respect to automatic metrics, and then with respect to human judgment via human-bot chats followed by the question {\em ``How much did you enjoy talking to this user?"} on a scale of 1--4. A total of 9 systems were evaluated using human annotators, 100 conversations for each. In this work, we leverage the human-model chat logs from the ConvAI2 competition for three models: Lost in Conversation ({\bf LIC})\footnote{\url{https://github.com/atselousov/transformer_chatbot}}, which won the competition, and Hugging Face ({\bf HF}; \citeauthor{wolf2019transfer}, \citeyear{wolf2019transfer}) which won the automatic evaluation track, and the KVMemNN \citep{miller2016keyvalue} baseline released by the competition organizers ({\bf KV}; \citeauthor{dinan2019second}, \citeyear{dinan2019second}). LIC and HF are large pretrained and fine-tuned generative Transformer models, while KV is a retrieval model with no pretraining.

Secondly, we also compare to recently published models from \citet{see2019goodconversation}. The authors studied the effects of controllable generation.
and showed that Repetition-controlled ({\bf RC}), Inquisitive ({\bf INQ}), and Interesting ({\bf INT}) models obtained the highest human Likert scores in their study, however their comparison to models from other studies is not direct. We thus compare to these models as well;
we use the human-model conversation logs from their work, 100 for each model.

Finally, we also compare to the Polyencoder model~({\bf PE}, \citeauthor{DBLP:journals/corr/abs-1905-01969}, \citeyear{DBLP:journals/corr/abs-1905-01969}), a recent state-of-the-art retrieval model. It is a type of large Transformer architecture pretrained on Reddit, which learns a small number of global features to represent the input so that retrieval can be computed efficiently.
As no conversation logs were provided in that work, we 
additionally collect human-model conversations for that model. 

Overall, we benchmark 7 models, and compare them to human~({\bf H}) performance in a number of different settings: with human-model and self-chat 
over three questions: engagingness, humamnness and interestingness. 

\subsection{Wizard of Wikipedia task}

\begin{figure}[t]
    \centering
    \includegraphics[width=0.47\textwidth]{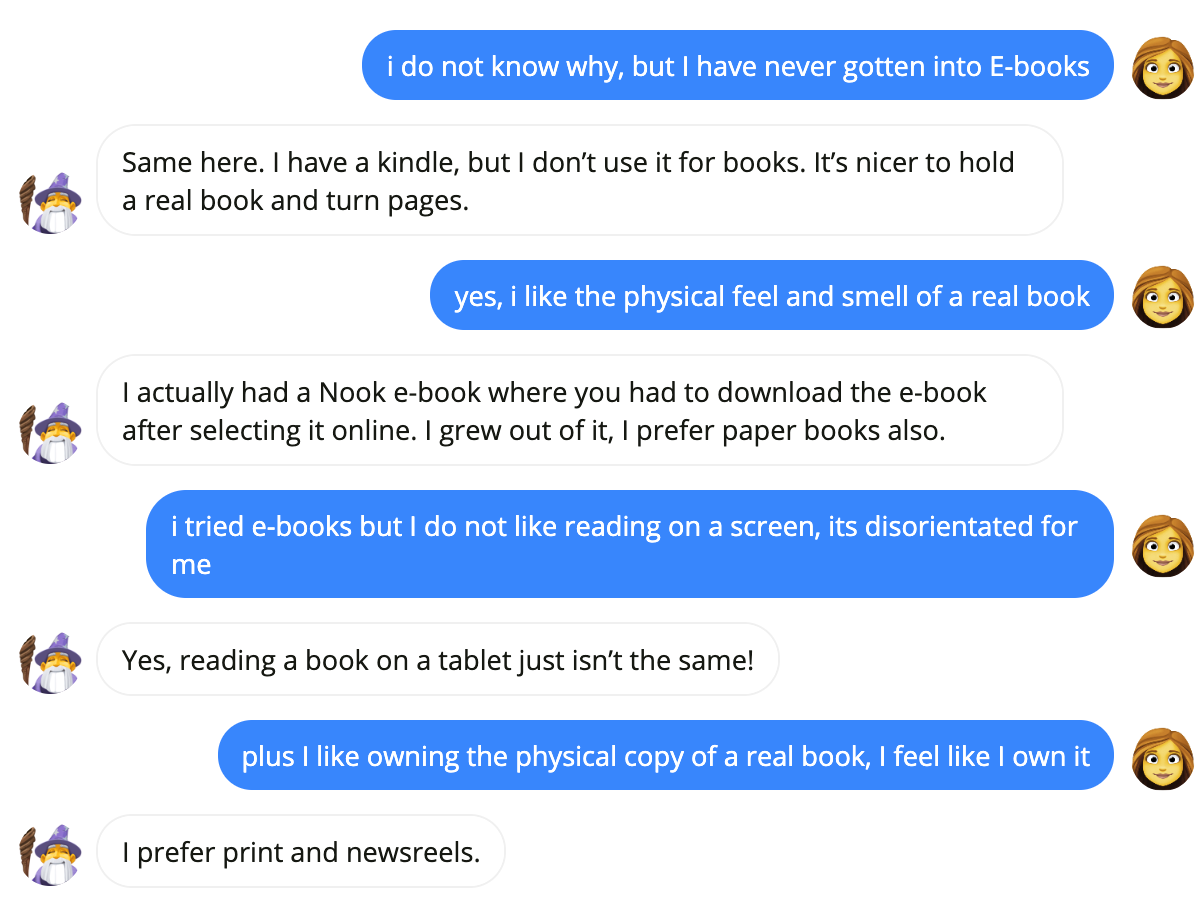}
    \caption{Example of the Wizard Retrieval (RK) talking with a human. The Wizard model is able to use facts from Wikipedia during its conversation.}
    \label{fig:wizex}
\end{figure}

Wizard of Wikipedia \cite{dinan2018wizard} is a chitchat dialogue task where two speakers discuss a topic in depth, chosen from 1247 topics. 
One speaker (termed the Wizard) is meant to be both engaging and knowledgeable on the topics, and has access to an information retrieval system over Wikipedia to supplement their own knowledge.
The other speaker (the Apprentice) is meant to be curious and eager to learn about the topic. The original dataset contains over 18,000 human-human dialogues, and has been used to train various kinds of models to imitate the human wizards.
These include the Memory Network Transformer, in both generative and retrieval versions that employs the retrieved knowledge by attending over it before producing an utterance ({\bf GK} and {\bf RK} respectively), and baselines that do not have access to the knowledge ({\bf GU} and {\bf RU}). See Figure~\ref{fig:wizex} for an example chat.
We use the human-model logs from that paper (100 conversations for each model) on unseen test topics and evaluate them against humans ({\bf H}),
using both engagingness and knowledgeability questions. 
We note the original paper tested engagingness only.

\subsection{Question Optimization}

We are interested in evaluating models in terms of four axes: engagingness, interestingness, knowledge and humanness.
In order to find the questions with highest inter-annotator agreement, we run multiple trials of experiments according to the setup described below. Each trial tests the effectiveness of a single question and consists of the same set of multi-turn conversation logs, presented to the human annotators. We test 13 questions: three regarding engagingness, four regarding interestingness, three regarding humanness, and three regarding knowledgeability (see Table~\ref{tab:abtestquestions}). 

We compare human-human logs with human-model logs where the model is a relatively low quality baseline model, with the aim that there should be a clear and agreeable difference between human and model which is visible to human annotators. For PersonaChat we use a greedy generative baseline, and for Wizard we use the GU (generative unknowledgeable) model. Both of these baselines exhibit strong repetitive behavior which is known to be highly disfavored by crowdworkers \cite{see2019goodconversation}. 
We select a single handpicked conversation pair for each of the tasks, and collect $\sim$20 annotations per question.

We calculate the inter-annotator agreement for each question. 
The question achieving the highest inter-annotator agreement is selected for use in the rest of our experiments. 
The specific question phrasing and the texts accompanying the option for Speaker 1 (i.e. the left-hand conversation) are listed in Table \ref{tab:abtestquestions} along with inter-annotator agreements.  As can be seen, the phrasing of the question is important,
with poor phrasing choices leading to much lower agreement levels, e.g. 86.7\% agreement in the best case for interestingness, and 69.6\% in the worst case.

As a preliminary sanity check, we ran A/A tests over each of the engagingness, interestingness, and humanness best questions, with the same model appearing as both Speaker 1 and 2. All three tests came back close to 50-50.

Overall, we  see this question optimization
step as an important pre-requisite for our main experiments, and use the best discovered phrasing in each case. We encourage
further research to use them as well.

\subsection{Benchmarking: Evaluation of State-of-the-art}
\label{sec:experiments}

\paragraph{PersonaChat}

\begin{table}[t!]

\setlength{\tabcolsep}{3pt}
    \centering
    \begin{small}
\begin{tabular}{rr|rrrrrrrr}

&  & \multicolumn{8}{c}{Wins \% matches}  \\
&        &   {RC}            &       {KV}      &   {INQ}         &        {HF}     &      {INT}     &       {LIC}     & {PE}            & {H}    \\
\midrule
\parbox[t]{2mm}{\multirow{9}{*}{\rotatebox[origin=c]{90}{Loses \% matches}}}
& RC     &                   & {\sffamily 50}          & \win{58}      & \win{54}      & \win{\textbf{66}} & \win{\textbf{68}}  & \win{\textbf{69}} & \win{\textbf {67}}\\[-0.25mm] 
& KV     &  {\sffamily 50}           &                 &  \win{57}     &  \win{55}     & \win{57}     & \win{57}      &\win{\textbf{61}}  & \win{\textbf {60}}\\[-0.25mm] 
& INQ    &  \lose{42}      & \lose{43}     &                 &  \win{51}     & \win{59}     & \win{52}      &\win{\textbf{62}}  & \win{\textbf{71}}\\[-0.25mm] 
& HF     &  \lose{46}      & \lose{45}     & \lose{49}     &                 & \win{55}     & \win{54}      &  \win{57}     & \win{\textbf{64}}\\[-0.25mm] 
& INT    &  \lose{\textbf{34}}  & \lose{43}     &  \lose{41}    &  \lose{45}    &                & \win{52}      &  \win{54}     & \win{52}    \\[-0.25mm] 
& LIC    &  \lose{\textbf{32}}  & \lose{43}     &  \lose{48}    &  \lose{46}    & \lose{48}    &                 &  \win{53}     & \win{\textbf {65}}\\[-0.25mm] 
& PE     & \lose{\textbf{31}}  & \lose{\textbf{39}}& \lose{\textbf{38}}&  \lose{43}    &  \lose{46}   & \lose{47}     &                 &\win{53}     \\[-0.25mm]
& H  &  \lose{\textbf{33}}  & \lose{\textbf{40}} & \lose{\textbf{29}} & \lose{\textbf{36}} & \lose{48}    & \lose{\textbf{35}} & \lose{47}     &               \\

\end{tabular}

    \end{small}
    \caption{{\sc{Acute-Eval}} results on the {\em Engagingness} question for the PersonaChat models talking to humans. Bold
    win percentages indicate significance ($p<.05$).}
    \label{tab:convairoundrobin}
\end{table}

\begin{table}[t]
\setlength{\tabcolsep}{3pt}
    \centering
    \begin{small}
\begin{tabular}{rr|rrrrrrrr}
&  & \multicolumn{8}{c}{Win Margin}  \\
    &        &        {RC}            &            {KV} &             {INQ}    &        {HF}      &              {INT}  &            {LIC} & {PE} &   {H}\\
\midrule
\parbox[t]{2mm}{\multirow{9}{*}{\rotatebox[origin=c]{90}{Lose Margin}}}
    & RC     &                       &                 &     \win{.18}        &                  &      \win{.10}      &                  &      & \win{\textbf{.42}}   \\[-0.25mm] 
    & KV     &                       &                 &                      &  \win{.17}      &                     &  \win{\textbf{.58}}   &      &                 \\[-0.25mm] 
    & INQ    &  \lose{-.18}          &                 &                      &                  &    \lose{-.08}      &                  &      & \win{\textbf{.24}}   \\[-0.25mm] 
    & HF     &                       &  \lose{-.17}    &                      &                  &                     &   \win{\textbf{.41}}  &      &                 \\[-0.25mm]
    & INT    &  \lose{-.10}          &                 &   \win{.08}          &                  &                     &                  &      & \win{\textbf{.32}}   \\[-0.25mm] 
    & LIC    &                       &\lose{\textbf{-.58}}  &                      & \lose{\textbf{-.41}}  &                     &                  &      &                 \\ 
    & PE     &                       &                 &                      &                  &                     &                  &      &                 \\[-0.25mm] 
    & H  &  \lose{\textbf{-.42}}      &                 &  \lose{\textbf{-.24}}     &                  &    \lose{\textbf{-.32}}  &                  &      &                 \\[-0.25mm]
\end{tabular}
    \end{small}
    \caption{Likert pairwise differences for {\em Engagingness} on PersonaChat, where known. Differences are collected from multiple papers and may not be directly comparable.}
    \label{tab:likert-convairoundrobin}
\end{table}

\begin{table}[t]
\setlength{\tabcolsep}{3pt}
    \centering
    \begin{small}
\begin{tabular}{rr|rrrrrrrr}
&  & \multicolumn{8}{c}{Wins \% matches}  \\
&         &          {RC}    &             {KV} &          {INQ}    &             {HF}   &      {INT}      &           {LIC} &          {PE}   & {H}       \\
\midrule
\parbox[t]{2mm}{\multirow{9}{*}{\rotatebox[origin=c]{90}{Loses \% matches}}}
& RC      &                  &     \win{58}   & \win{\textbf{67}}    & \lose{\textbf{42}}      & \win{\textbf{73}}  & \win{\textbf{68}}  & \win{\textbf{74}}  & \win{\textbf{74}}\\[-0.25mm]
& KV      & \lose{42}        &                  & \win{51}        & \lose{\textbf{26}}    & \win{57}      & \win{\textbf{60}}  & \win{\textbf{63}}  & \win{\textbf{71}}\\[-0.25mm]
& INQ     & \lose{\textbf{33}}    & \lose{49}      &                   & \lose{\textbf{25}}    & \win{\textbf{63}}  & \win{\textbf{66}}  & \win{\textbf{63}}  & \win{\textbf{72}}\\[-0.25mm]
& HF      & \win{\textbf{58}}     & \win{\textbf{74}}   & \win{\textbf{75}}    &                    & \win{\textbf{81}}  & \win{\textbf{81}}  & \win{\textbf{82}}  & \win{\textbf{81}}\\[-0.25mm]
& INT     & \lose{\textbf{27}}    & \lose{43}      & \lose{\textbf{37}}   & \lose{\textbf{16}}      &               & \win{51}      & \win{51}      & \win{\textbf{63}}\\[-0.25mm]
& LIC     & \lose{\textbf{32}}    & \lose{\textbf{40}}  & \lose{\textbf{34}}   & \lose{\textbf{19}}      & \lose{49}     &               & \win{55}      & \win{\textbf{60}}\\[-0.25mm]
& PE      & \lose{\textbf{26}}    & \lose{\textbf{37}}  & \lose{\textbf{37}}   & \lose{\textbf{18}}      & \lose{49}     & \lose{45}     &               & \win{\textbf{61}}\\[-0.25mm] 
& H   &  \lose{\textbf{26}}   & \lose{\textbf{29}}  & \lose{\textbf{28}}   & \lose{\textbf{19}}      & \lose{\textbf{37}} & \lose{\textbf{40}} & \lose{\textbf{39}} &             \\
\end{tabular}
    \end{small}
    \caption{{\sc{Acute-Eval}} results for self-chats   for the {\em Engagingness} question on PersonaChat. Results largely agree with the human-model evaluations (Table~\ref{tab:convairoundrobin}) and the Likert evaluations (Table~\ref{tab:likert-convairoundrobin}).} 
    \label{tab:selfchatroundrobin}
\end{table}

We first compare all 7 models and humans on the PersonaChat task using {\sc Acute-eval} over the human-model chats using the optimized engagingness question. In total, we evaluate 28 paired comparisons. Results are given in Table~\ref{tab:convairoundrobin}. 
Bold win percentages indicate significance.

We first observe that the models form a clean well-ordered set, and there are no rock-paper-scissors effects, giving an order Human~$>$~PE~$>$~LIC~$>$~INT~$>$~HF~$>$~INQ~$>$~KV~$>$~RC. In general, these results agree closely with the known Likert comparisons made in prior papers, shown in Table~\ref{tab:likert-convairoundrobin}.
Similar conclusions are derived for the interestingness and humanness questions as
well, see Tables~\ref{tab:convairoundrobininterestingness} and \ref{tab:convairoundrobinhumanness}, note the model ordering is slightly different for those questions. \citet{see2019goodconversation} previously showed that different models often exhibit different rankings for different metrics, and {\sc Acute-eval} results remain largely consistent with Likert.
 
A surprising result for the community is that the retrieval model PE outperforms 
all generative models, as the community has focused heavily on building 
generative models, e.g. almost all 23 entrants to the ConvAI2 competition \cite{dinan2019second}.
Now that the current best performing 
models have been benchmarked against each other we hope future research
will use the same approach so the state-of-the-art can be clearly tracked.

\paragraph{Self-Chat}

\begin{figure}[t]
    \centering
    \includegraphics[width=0.47\textwidth]{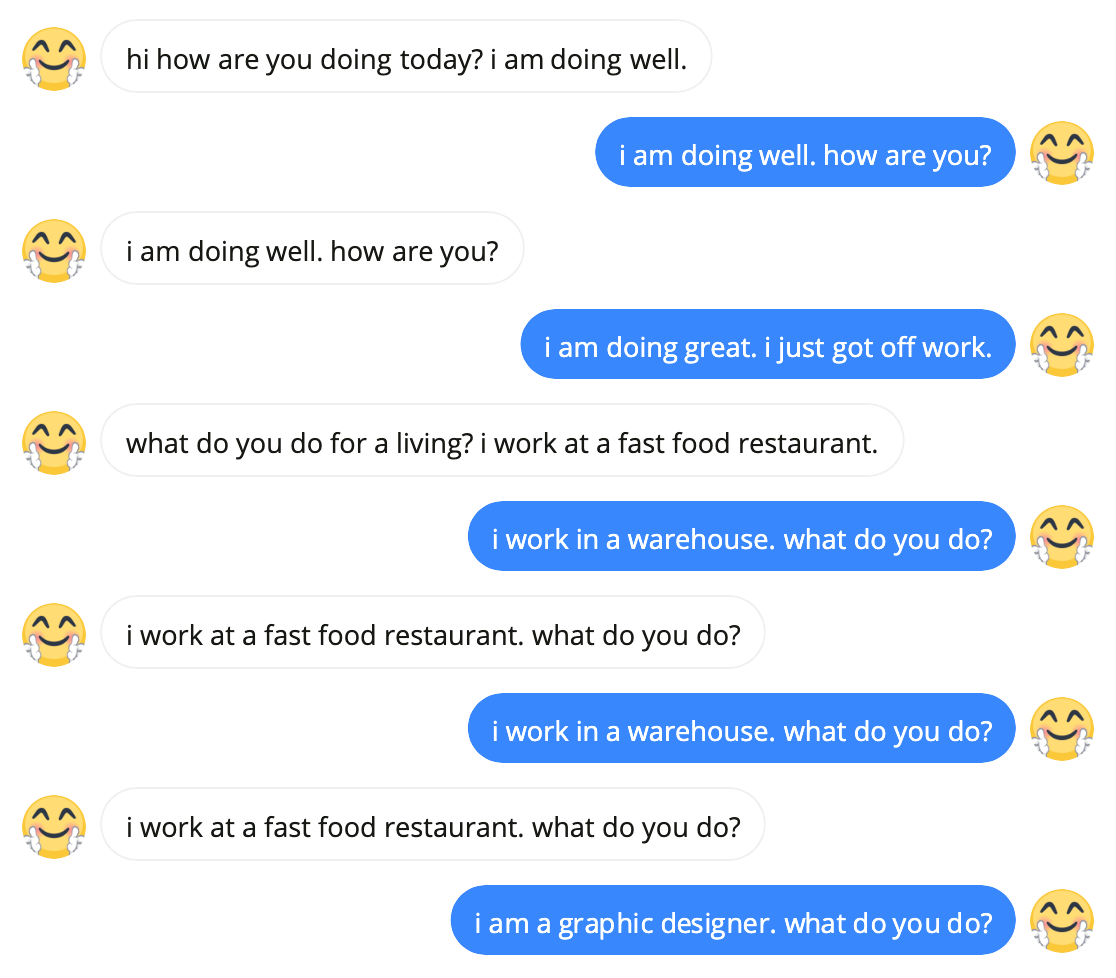}
    \caption{Randomly chosen example of Hugging Face (HF) model talking with itself. HF self-chat degenerates rapidly, explaining its poor performance. Other models handle self-chat more successfully, see  Fig.~\ref{fig:peself} and Supplementary Material.}
    \label{fig:hfself}
\end{figure}

\begin{figure}[t]
    \centering
    \includegraphics[width=0.44\textwidth]{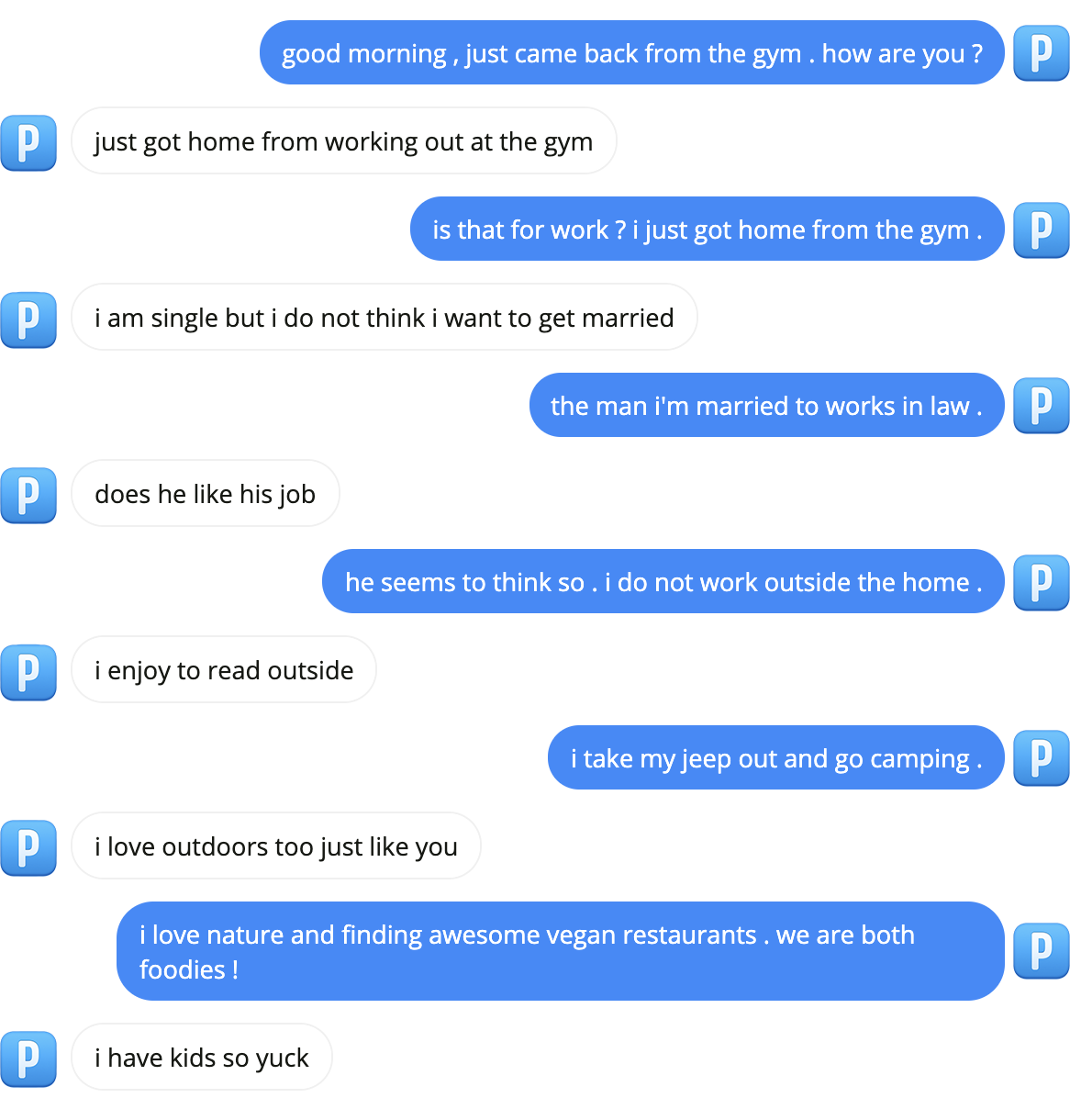}
    \caption{Randomly chosen example of Polyencoder (PE) model talking with itself (self-chat). }
    \label{fig:peself}
\end{figure}

\begin{table}[t]
\setlength{\tabcolsep}{3pt}
    \centering
    \begin{small}
\resizebox{\linewidth}{!}{
\begin{tabular}{r|rrrrr|rrrrr}
& \multicolumn{5}{c|}{Wins \%}  & \multicolumn{5}{c}{Win Margin}\\
       &       {RC}    & {LIC}         &       {INT}  & {PE}         & {H}         &       {RC}     & {LIC}&        {INT}   & {PE} & {H}             \\
\midrule
RC     &               & \win{53}      & \win{\textbf{64}} & \win{\textbf{68}} & \win{\textbf{73}}&                &      &\lose{-.01}     &      &\win{\textbf{.90}}    \\[-0.25mm]
LIC    & \lose{47}     &               & \win{54}     & \win{56}     & \win{\textbf{59}}&                &      &                &      &                 \\[-0.25mm] 
INT    & \lose{\textbf{36}} & \lose{46}     &              & \win{51}     & \win{\textbf{59}}&\win{-.01}     &      &                &      &\win{\textbf{.91}}    \\[-0.25mm] 
PE     & \lose{\textbf{32}} & \lose{44}     & \lose{49}    &              & \win{54}    &                &      &                &      &                 \\[-0.25mm]
H      & \lose{\textbf{27}} & \lose{\textbf{41}} & \lose{\textbf{41}}& \lose{46}    &             &\lose{\textbf{-.90}} &      &\lose{\textbf{-.91}} &      &                 \\
\end{tabular}
}
    \end{small}
    \caption{Results on the {\em Humanness} question for the PersonaChat models talking to humans.
    {\sc{Acute-Eval}} (left)  is able to identify significant differences between INT and RC when Likert (known published differences, right) does not.
    }
    \label{tab:convairoundrobinhumanness}
\end{table}

We perform {\sc Acute-eval} over self-chats instead of human-model chats.
We compare all models and humans (via human-human chats) in an otherwise 
identical setup to the human-bot evaluation for PersonaChat.
Results are given in Table \ref{tab:selfchatroundrobin}.

We observe very similar conclusions to human-model chats in terms of winning models, making this a viable cheaper alternative to collecting human-model conversations, thus being considerably cheaper to collect. This approach also appears to require relatively fewer annotations/person-hours in this case to achieve statistical significance.
One important caveat is the performance of the HF model.  HF self-chats surface degeneracies in the model itself, and do not look natural (see Figure~\ref{fig:hfself} for examples), explaining its poor performance compared to all other models. 
All other models do not exhibit this behavior and apart from HF, 
are ordered by humans exactly the same as for human-bot chats. 
For example, see Figure~\ref{fig:peself} for PE engaging in self-chat more successfully.
However, due to the inadequacies of a specific model, in this case HF, conclusions from self-chat performance results must therefore be handled with care,
but we believe are a reasonable choice for early experiments in the model development cycle,
enabling faster research iteration.

One concern with self-chat is that powerful models could easily cheat, and simply recall training examples with perfect accuracy. In practice, we found that none of the models exhibit this behavior: $<$1\% of the Polyencoder's call-response utterance pairs produced during self-chats come directly from the training set. The worst offender, INQ, has roughly 10\% of pairs coming from training, but this stems from it using the same generic greeting and response in nearly all conversations (``Hello, how are you doing today?'', ``I am doing well, how about yourself?'').

\begin{table}[t]
\setlength{\tabcolsep}{3pt}
    \centering
    \begin{small}
\resizebox{\linewidth}{!}{
\begin{tabular}{r|rrrrr|rrrrr}
& \multicolumn{5}{c|}{Wins \%}  & \multicolumn{5}{c}{Win Margin}\\
        &{RC}          &{LIC}          &       {INT}   & {PE}         & {H}          &{RC}            &{LIC} &       {INT}    & {PE} & {H} \\
\midrule
RC     &               & \win{52}      & \win{\textbf{71}}  & \win{\textbf{75}} & \win{\textbf{76}} &                &      &\win{.04}      &      &\win{\textbf{.26}}\\[-0.25mm]
LIC    & \lose{48}     &               & \win{57}      & \win{\textbf{66}} & \win{\textbf{66}} &                &      &                &      &             \\[-0.25mm]
INT    & \lose{\textbf{29}} & \lose{43}     &               & \win{55}     & \win{\textbf{64}} &\lose{-.04}    &      &                &      &\win{\textbf{.23}}\\[-0.25mm]
PE     & \lose{\textbf{25}} & \lose{\textbf{34}} & \lose{45} &              & \win{52}     &                &      &                &      &             \\[-0.25mm]
H      & \lose{\textbf{24}} & \lose{\textbf{34}} & \lose{\textbf{36}} & \lose{48}    &              &\lose{\textbf{-.26}} &      &\lose{\textbf{-.23}} &      &             \\
\end{tabular}
}
    \end{small}
    \caption{Results on the {\em Interestingness} question for the PersonaChat models talking to humans. 
    {\sc{Acute-Eval}} (left)  is able to identify significant differences between INT and RC when Likert (known published differences, right) does not.
    }
    \label{tab:convairoundrobininterestingness}
\end{table}

\begin{table}[t]
\setlength{\tabcolsep}{2.5pt}
    \centering
    \begin{small}
\resizebox{\linewidth}{!}{
\begin{tabular}{r|rrrrr|rrrrr}
& \multicolumn{5}{c|}{Wins \%}  & \multicolumn{5}{c}{Win Margin}\\
         & {GU}            & {GK}            & {RU}           & {RK}            & {H}  & {GU}            & {GK}            & {RU}           & {RK}            & {H}\\
\midrule
GU &               &  \win{\textbf{67}} & \win{\textbf{79}} & \win{\textbf{75}}  & \win{\textbf{77}}  &                &\win{.39}       &\win{\textbf{.58}}   &\win{\textbf{.60}}   &\win{\textbf{1.8}} \\[-0.25mm]
GK & \lose{\textbf{33}} &               & \win{\textbf{64}} & \win{\textbf{63}}  & \win{\textbf{73}}  &\lose{-.39}     &                &\win{.19}       &\win{.21}       &\win{\textbf{1.4}} \\[-0.25mm]
RU & \lose{\textbf{21}} & \lose{\textbf{36}} &              & \win{52}      &     \lose{48} &\lose{\textbf{-.58}} &\lose{-.19}     &                &\win{.02}       &\win{\textbf{1.2}} \\[-0.25mm]
RK & \lose{\textbf{25}} & \lose{\textbf{37}} & \lose{48}    &               & \win{\textbf{62}}  &\lose{\textbf{-.60}} &\lose{-.21}     &\lose{-.02}     &                &\win{\textbf{1.2}} \\[-0.25mm]
H  & \lose{\textbf{23}} & \lose{\textbf{27}} & \win{52}     & \lose{\textbf{38}} &               &\lose{\textbf{-1.8}} &\lose{\textbf{-1.4}} &\lose{\textbf{-1.2}} &\lose{\textbf{-1.2}} &              \\
\end{tabular}
}
    \end{small}
    \caption{Results on the {\em Engagingness} question for the Wizard of Wikipedia models (G/R for Generative/Retrieval and U/K for with and without access to knowledge.
    Left shows the {\sc{Acute-Eval}} results, and right shows known Likert differences. Our method shows statistical significance between several methods that Likert does not.}
    \label{tab:wizardengage}
\end{table}

\begin{table}[t]
\setlength{\tabcolsep}{3pt}
    \centering
    \begin{small}
\begin{tabular}{rr|rrrrr}
&  & \multicolumn{5}{c}{Wins \%}  \\
&         & {GU}    & {GK}          & {RU}       & {RK}       & {H}\\
\midrule
\parbox[t]{2mm}{\multirow{5}{*}{\rotatebox[origin=c]{90}{Loses \%}}}
& {GU}    &                 &\win{\textbf{79}}   &\win{\textbf{85}} &\win{\textbf{82}}  &\win{\textbf{76}} \\[-0.25mm]
& {GK}    &\lose{\textbf{21}}    &               &\win{54}     &\win{\textbf{70}}  &\win{56}     \\[-0.25mm]
& {RU}    &\lose{\textbf{15}}    &\lose{46}      &             &\lose{49}     &\lose{48}    \\[-0.25mm]
& {RK}    &\lose{\textbf{18}}    & \lose{\textbf{30}} &\win{51}     &              &\lose{47}     \\[-0.25mm]
& {H}     &\lose{\textbf{24}}    &\lose{44}      &\win{52}     &\win{53}     &             \\
\end{tabular}
    \end{small}
    \caption{{\sc{Acute-Eval}} results on the {\em Knowledgeability} question for Wizard of Wikipedia models (G/R for Generative/Retrieval and U/K with and without access to knowledge.}
    \label{tab:wizardknowledge}
\end{table}

\paragraph{Wizard of Wikipedia}
We similarly compare all 4 models and humans on the optimized 
engaging and knowledge questions. The results are given in Tables~\ref{tab:wizardengage} and \ref{tab:wizardknowledge}.
We again find retrieval models outperform generative models, with knowledge attention (GK)
clearly helping the generative models, but with RU and RK very close.

Results largely agree between the two questions,
except retrieval with knowledge (RK) more clearly beats the generative version (GK) than retrieval without (RU)  when the question is about knowledge. For the engagingness question, where it makes sense that this is less important, there is little difference between knowledge or not.

\begin{figure}[t]
    \centering
    \includegraphics[width=0.47\textwidth]{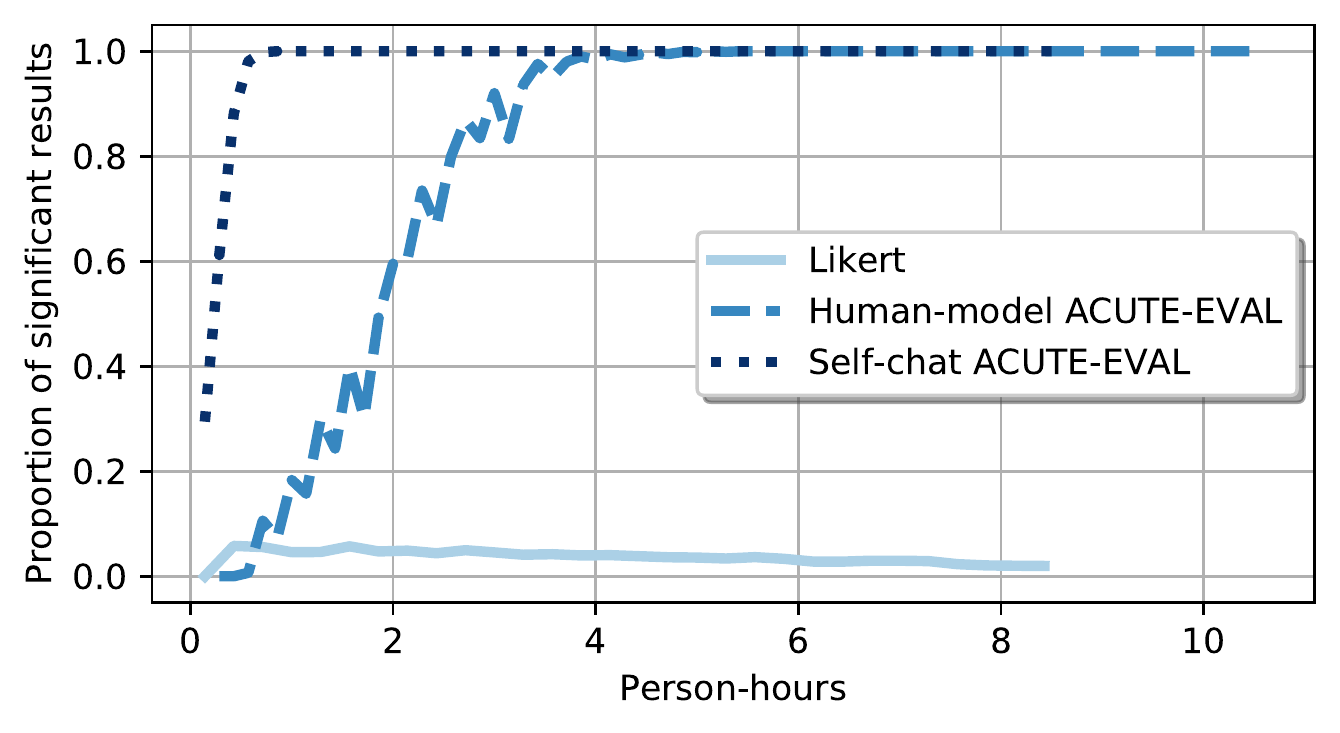}
    \caption{Relative cost effectiveness of potential collection methods: Likert and {\sc Acute-eval} human-model chat and self-chat pairwise tests. Our methods obtain statistical significance with fewer person hours; Likert fails in this case.
    }
    \label{fig:cost_graph}
\end{figure}

\paragraph{Comparison to Likert}
We compare {\sc Acute-eval} to multi-turn Likert for both tasks by computing
pairwise Likert differences, where known, from the original papers. We do not compare across papers as evaluation setups differ.
Values are provided  in Tables~\ref{tab:likert-convairoundrobin},
\ref{tab:convairoundrobininterestingness},
\ref{tab:convairoundrobinhumanness}
and \ref{tab:wizardengage}.
While the tests generally agree,
{\sc Acute-eval} can be a more sensitive test, which 
more often yields significance. 
On Wizard of Wikipedia where all Likert matchups are known,
8 of the pairwise matchups are significant for our test with human-model chats, while 6 are significant for Likert. 
On PersonaChat for the interestingness question, 6 of 10 matchups are significant for {\sc Acute-eval}, including all known  Likert matchups, which only has 2 of 3 that are significant.
For the humanness question, 5 of 10 matchups are significant for {\sc Acute-eval}, including all known  Likert matchups, which only has 2 of 3 that are significant.
For the engagingness question, 5 of the 9 Likert matchups are significant.
All 9 are significant for {\sc Acute-eval} when using self-chats; 
3 are significant for human-model chats.

We compare the cost effectiveness of Likert to {\sc Acute-eval} human-model and self-chat comparisons in 
Figure~\ref{fig:cost_graph}. Shown is the PersonaChat {\em Engagingness} question comparing RC and INT models, 
a fairly tight matchup. We show the \% chance of achieving
significance when drawing pairs of dialogues at random, plotting with respect to person-hours spent annotating. 
In this case Likert fails to achieve significance, likely due
to bias and variance issues with integer scores. {\sc Acute-eval} human-model and self-chat pairwise tests perform well, achieving
significance; self-chat requires fewer person-hours.

\section{Conclusion}
\label{sec:conclusion}

Studying the ability of machines to communicate with humans is an important long-term
goal of AI research. Unfortunately, measuring progress towards that goal has been hampered
by the trustworthiness of evaluation itself. Current human evaluation methods such 
as multi-turn Likert 
are expensive to run, have annotator bias and variance problems, and can fail to yield statistical significance. 

In this work we have contributed
a novel evaluation method that alleviates some of these problems.
By optimizing questions and performing comparisons on pairs of human-bot dialogues we
arrive at more sensitive statistical tests when benchmarking current state-of-the models.
Utilizing self-chat bot evaluations we can often
improve sensitivity, while yielding even cheaper evaluations.
We will publicly release the code for our tests, and recommend them to
be used in future research studies in order to push forward the state of the art.

\clearpage
\bibliography{main}
\bibliographystyle{aaai}
\clearpage

\onecolumn

\section*{Supplementary Material}

\begin{figure}[!h]
    \centering
    \includegraphics[width=0.47\textwidth]{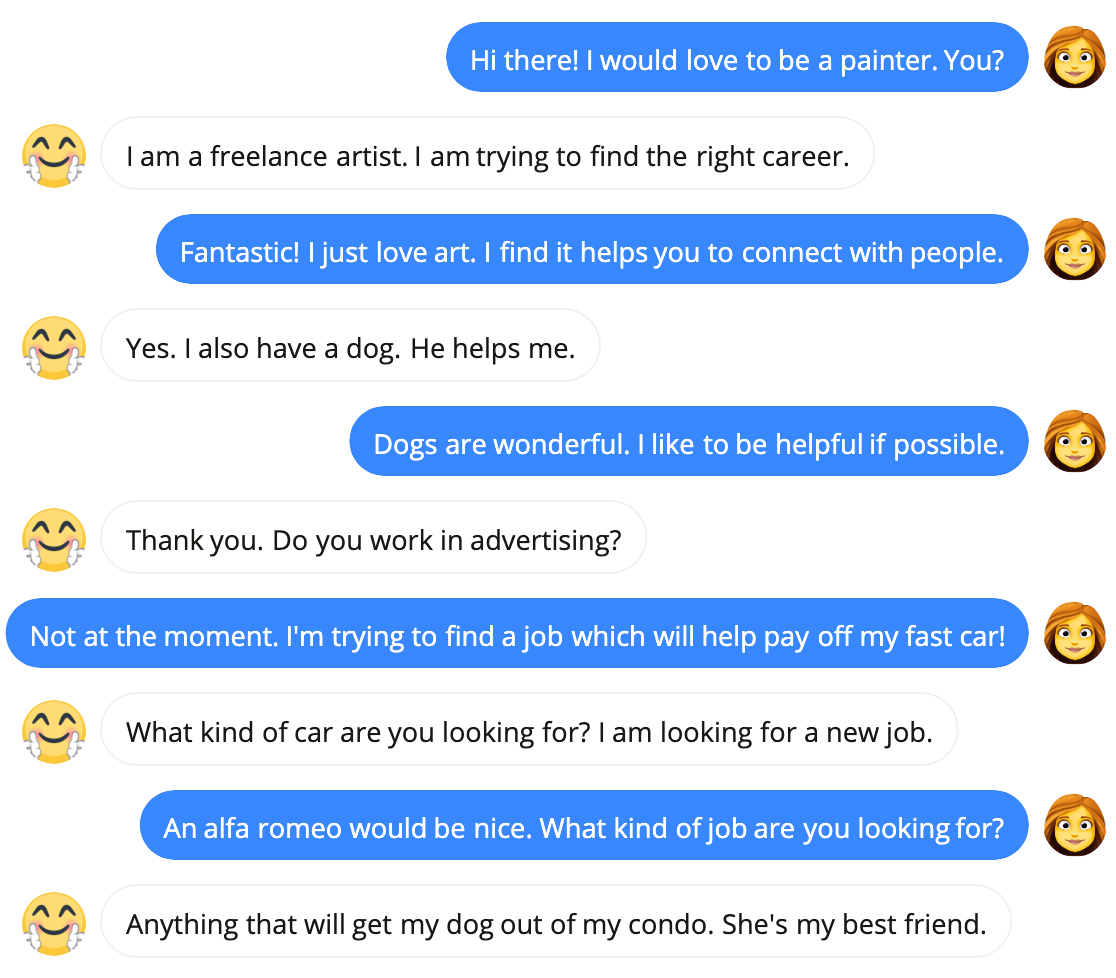}
    \quad
    \includegraphics[width=0.47\textwidth]{hf_selfchat.png}
    \caption{Randomly chosen examples of Hugging Face (HF) model talking with with a human (left) and itself (self-chat, right). HF self-chat degenerates rapidly, explaining its poor performance. Other models do not have this degeneration feature.}
    \label{fig:hfhumanself}
\end{figure}

\begin{figure}[!h]
    \centering
    \includegraphics[width=0.47\textwidth]{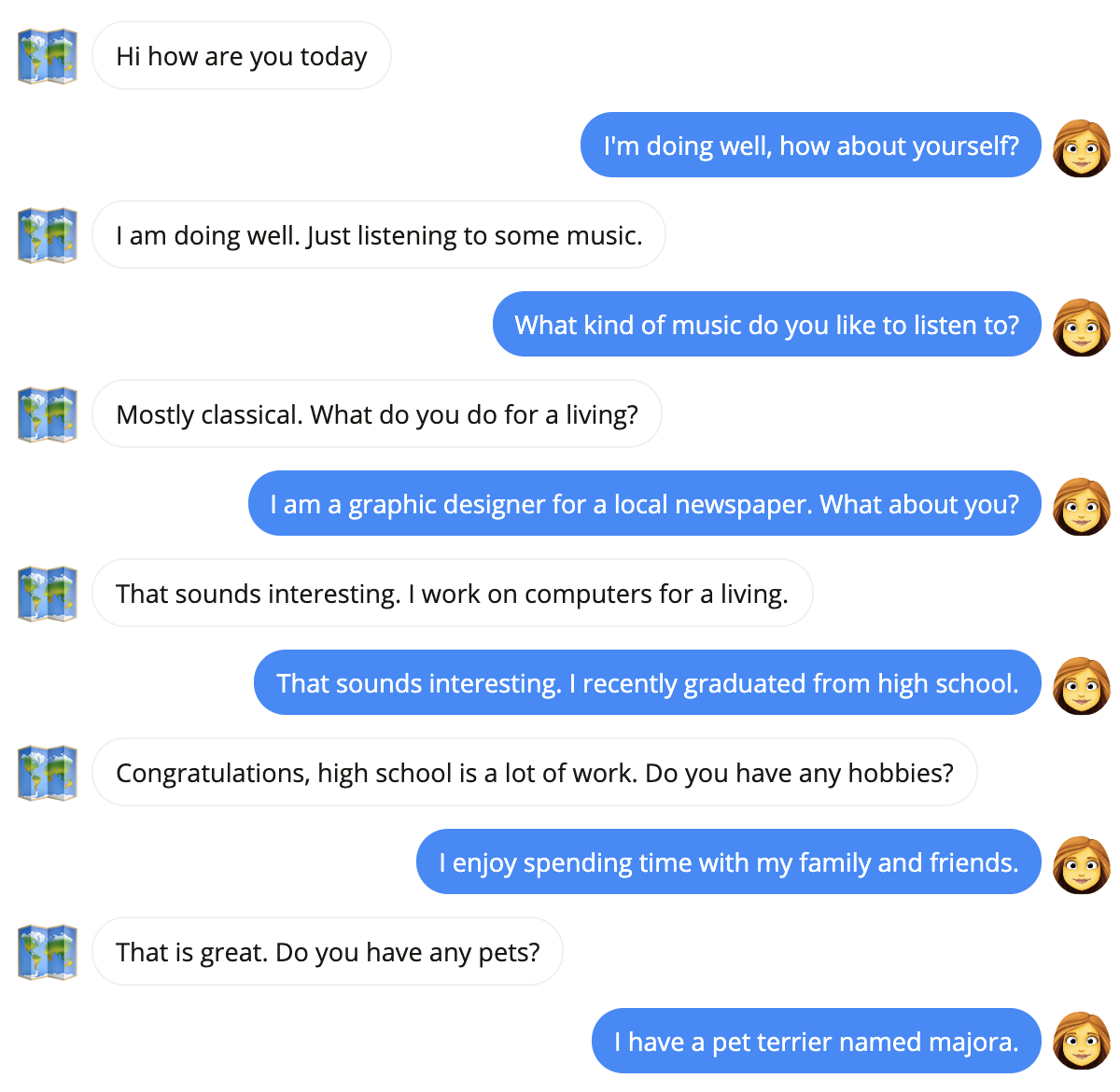}
    \quad
    \includegraphics[width=0.47\textwidth]{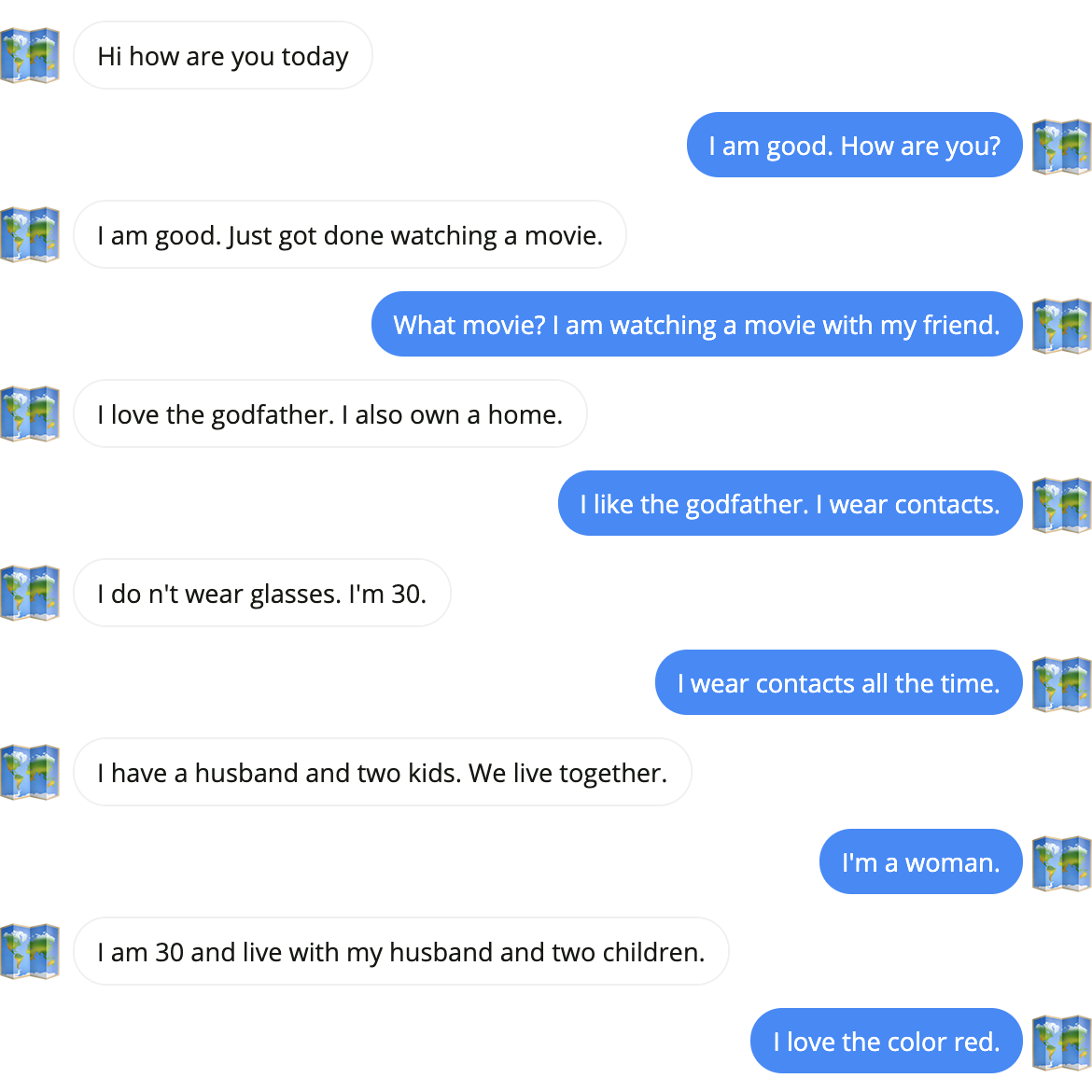}
    \caption{Examples of Lost in Conversation (LIC) model talking with a human subject (left), and itself (right). Both examples were selected randomly.}
    \label{fig:licselfvshuman}
\end{figure}

\begin{figure}[!h]
    \centering
    \includegraphics[width=0.47\textwidth]{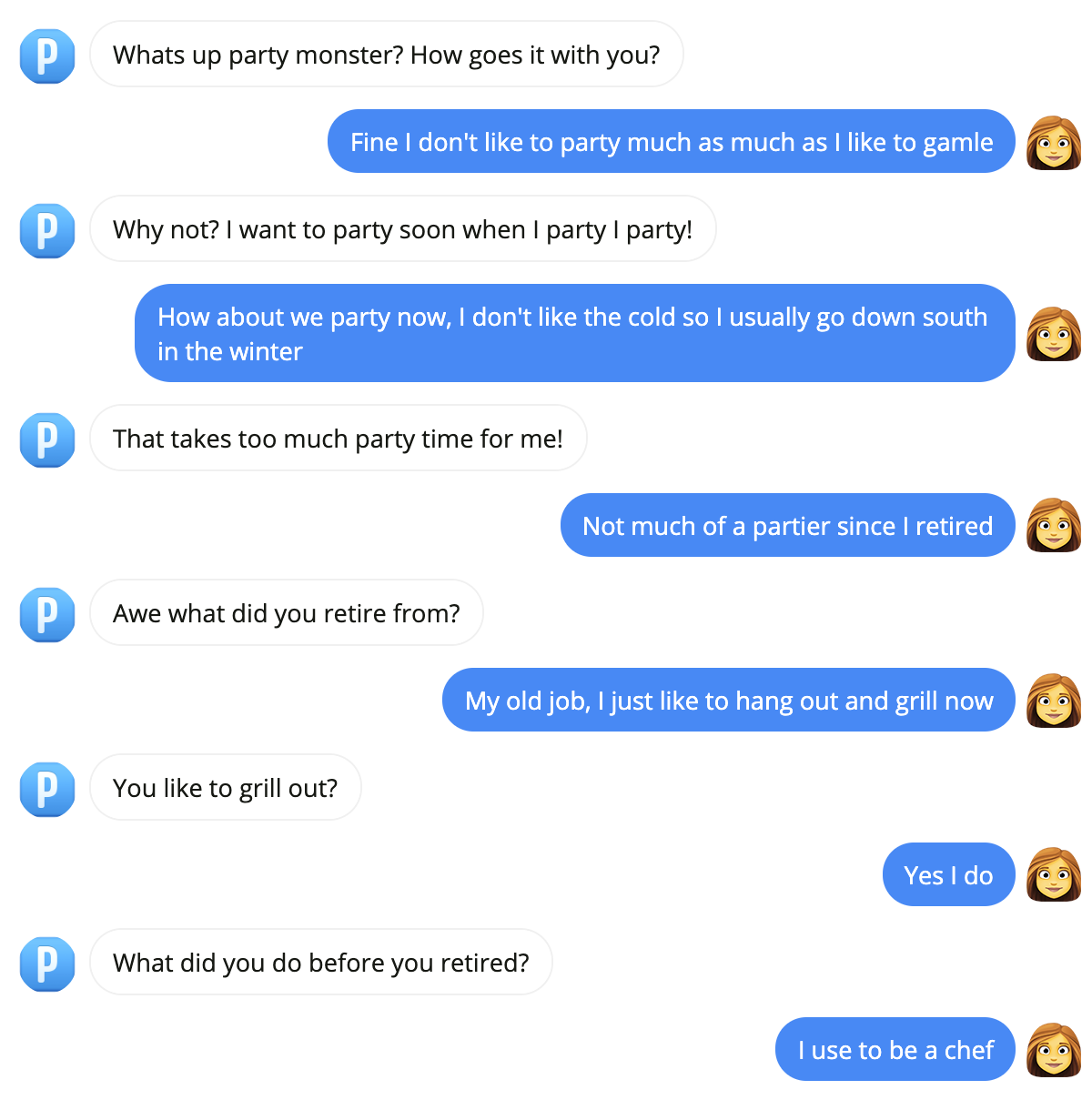}
    \quad
    \includegraphics[width=0.47\textwidth]{pe_selfchat.png}
    \caption{Examples of Polyencoder (PE) model talking with a human subject (left), and itself (right). Both examples were selected randomly. }
    \label{fig:peselfvshuman}
\end{figure}

\begin{figure}[!h]
    \centering
    \includegraphics[width=0.47\textwidth]{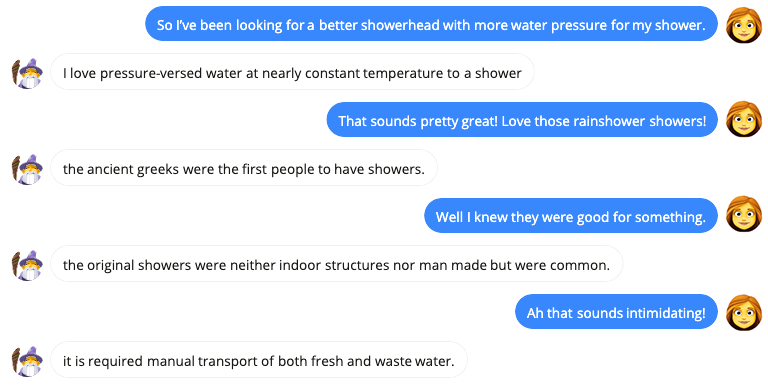}
    \quad
    \includegraphics[width=0.47\textwidth]{wizrkhuman.png}
    \caption{Examples of Wizard of Wikipedia chats. Left shows Generative model (GK) talking with a human subject. Right shows the Retrieval model (RK).}
    \label{fig:wiz2}
\end{figure}

\end{document}